\documentclass[lettersize,journal]{IEEEtran}
\usepackage{amsmath,amsfonts}
\usepackage{algorithmic}
\usepackage{array}
\usepackage[caption=false,font=normalsize,labelfont=sf,textfont=sf]{subfig}
\usepackage{textcomp}
\usepackage{stfloats}
\usepackage{url}
\usepackage{verbatim}
\usepackage{graphicx}
\usepackage{amssymb} 
\usepackage{tablefootnote}
\usepackage{tabularx} 
\usepackage{booktabs}
\usepackage{multirow}
\usepackage{hyperref}

\def\BibTeX{{\rm B\kern-.05em{\sc i\kern-.025em b}\kern-.08em
    T\kern-.1667em\lower.7ex\hbox{E}\kern-.125emX}}
\usepackage{balance}

\begin{document}
\title{Bridging Thought and Action: Taming Long-Horizon Instability in Open-Source LLM Agents with a MetaTool-Enhanced ROS Framework}

\author{
Kazi Abrar Mahmud, 
Nilotpaul Kundu Dhurubo, 
Tamal Kirttonia, 
Sabbir Hossain Ujjal
and Mohammad Ariful Haque%
\thanks{Kazi Abrar Mahmud is the corresponding author 
(email: abrar.mahmud790011@gmail.com).}%
\thanks{K. A. Mahmud, N. K. Dhurubo, T. Kirttonia, and M. A. Haque are with the Department of Electrical and Electronic Engineering, Bangladesh University of Engineering and Technology (BUET), Dhaka 1000, Bangladesh 
(emails: abrar.mahmud790011@gmail.com; nilbuet60@gmail.com; tamalkirttaniaofficial@gmail.com; arifulhoque@eee.buet.ac.bd).}%
\thanks{S. H. Ujjal is with the University of Massachusetts Amherst, 300 Massachusetts Ave, Amherst, MA 01003, United States 
(email: sujjal@umass.edu).}%
\thanks{This work has been submitted to the IEEE for possible publication. Copyright may be transferred without notice, after which this version may no longer be accessible.}%
}


\maketitle

\begin{abstract}
Large Language Models (LLMs) have enabled more natural human–robot interaction,
but open-source models often exhibit unstable long-horizon reasoning and inefficient action execution when deployed in agentic robotic frameworks. This paper presents an enhanced ROS-Agent based architecture that improves task reliability and execution efficiency for agentic robotic systems using open-source LLMs. The proposed system introduces a novel intermediate mechanism, termed the \emph{MetaTool}, which enforces structured planning prior to action execution. Given a natural-language command, the MetaTool induces the LLM to generate a pseudo-code plan of intended tool invocations, which is stored in the ROS-Agent's scratchpad and persists throughout execution. By explicitly separating planning from execution, the proposed approach reduces execution loops and improves deterministic behavior. The architecture is validated on a custom mobile robotic platform with multimodal perception and motion control capabilities. Experimental results on real-world interactive tasks demonstrate improved task completion and contextual consistency, with up to $\sim 24\%$ gains on complex tasks compared to the baseline framework.
\end{abstract}


\section{Introduction}
\IEEEPARstart{R}{ecent} advances in large language models (LLMs) have enabled
natural language–based robotic interaction, allowing robots to interpret and execute
human instructions through integrated reasoning and tool use. Agentic
frameworks that couple LLMs with robotic middleware have emerged as a promising
approach for such interaction \cite{10730448,10891883,10610981}. In particular, the Robot Operating System Agent
(ROSA) framework \cite{royce2025enabling}, developed by NASA JPL, provides a
structured interface for integrating LLM-based reasoning with ROS-native tools,
supporting task decomposition and execution within robotic systems \cite{Quigley2009ROSAO,6174325}.

However, deploying open-source LLMs in agentic robotic pipelines remains
challenging. Long-horizon reasoning across multiple tool invocations often
leads to instability, including redundant actions, execution loops, and
inconsistent task progression \cite{Liu2023LostIT,10610065}. These issues limit execution reliability for
complex, multi-step tasks. While proprietary models generally exhibit more stable behavior, their computational cost and limited accessibility motivate
architectural solutions that improve the robustness of open-source alternatives.
\begin{figure}[t]
    \centering
    \includegraphics[width=0.48\textwidth]{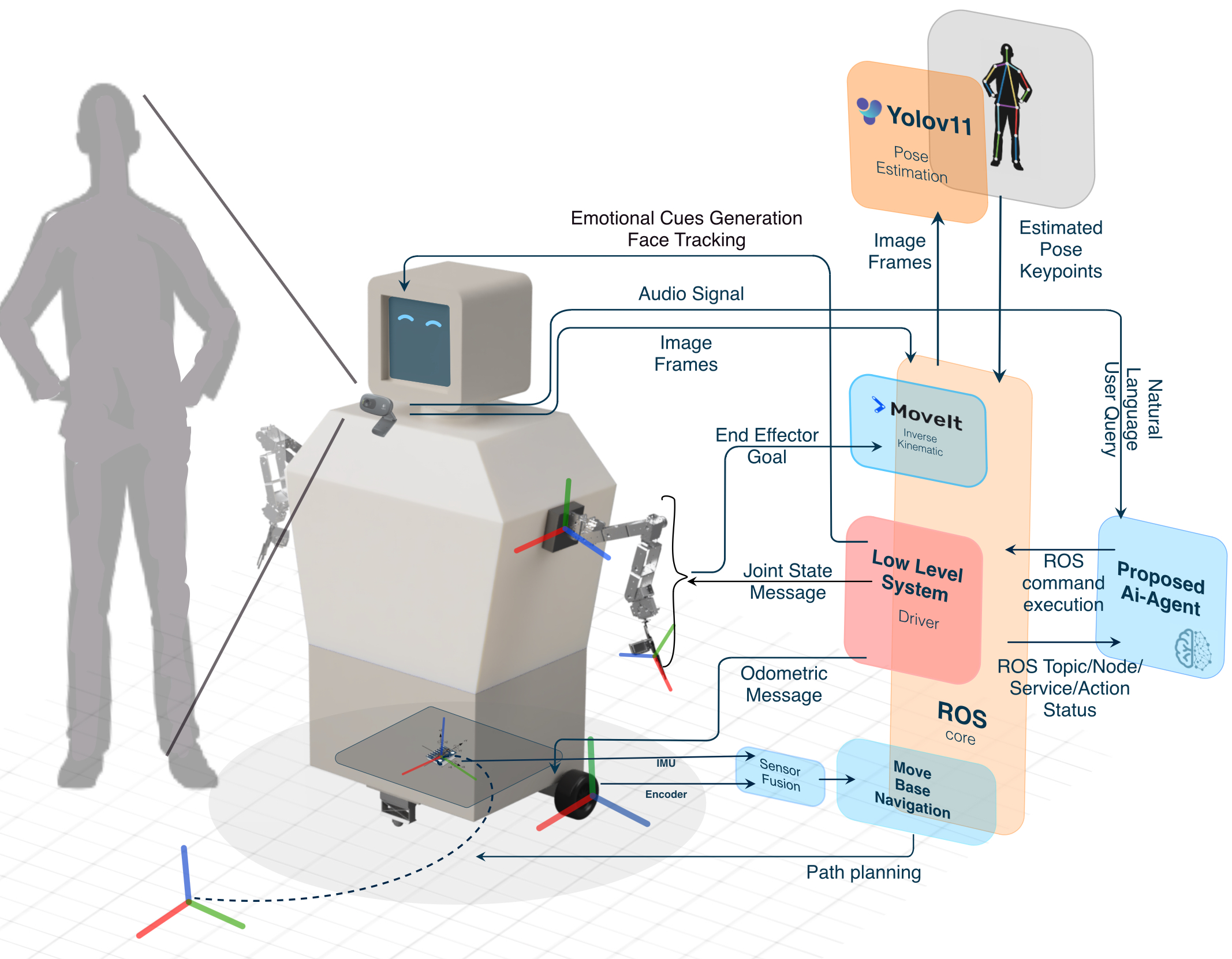}
\caption{Overview of the custom-designed robotic platform developed for system evaluation. The platform integrates perception modules, facial emotion display, articulated arm gestures, and mobility components, all managed through a ROS-based control stack.}

    \label{overall_archi}
\end{figure}
To address these challenges, we propose an enhanced ROS-Agent-based architecture
that introduces a novel intermediate mechanism termed the \emph{MetaTool}. The MetaTool operates between the LLM and the ROS execution layer, inducing the model to generate a structured pseudo-code representation of the intended tool sequence prior to execution. This plan is stored in ROS-Agents’s scratchpad memory and persists throughout the observation–reasoning–action loop, explicitly conditioning subsequent tool invocations. The ROS-Agent framework then invokes the selected tools to execute the corresponding robotic actions.

By externalizing task intent into a persistent execution plan, the MetaTool
constrains reasoning and action selection, mitigating common failure modes in
prompt-driven agentic systems and promoting more predictable execution behavior.
We evaluate the proposed architecture through systematic ablation and task-level experiments across multiple models and task complexities, demonstrating consistent improvements in execution reliability, particularly for open-source LLMs. The framework is validated in both ROS–Gazebo simulations \cite{1389727} and on a custom-built mobile robotic platform, shown in  Fig.~\ref{overall_archi}, integrating perception modules, facial emotion display, articulated arm gestures, and mobility within a ROS-based control stack.

Table~\ref{tab:comparison} compares representative LLM-based robotic planning systems. Unlike prior approaches, the proposed MetaTool ROS-Agent framework combines explicit planning, execution constraints, ROS-native integration, and an open-LLM focus, validated on a real robotic platform.

The remainder of this paper presents related work, methodology, evaluation, and conclusions.


\begin{table*}[!t]
\caption{Feature comparison of LLM-based robotic planning systems.}
\label{tab:comparison}
\centering
\begin{tabularx}{\textwidth}{@{}l X X X X X@{}}
\toprule
System & Explicit Plan Buffer & Execution Constraint & ROS-native & Open-LLM Focus & Real Robot \\ \midrule
ReAct \cite{yao2022react}          & $\times$ & $\times$ & $\times$ & $\times$ & $\times$ \\
SayPlan \cite{pmlr-v229-rana23a}   & $\checkmark$ & $\checkmark$ & $\checkmark$ & $\times$ & $\checkmark$ \\
RoboGPT \cite{10891883}         & $\checkmark$ & $\sim$ & $\times$ & $\checkmark$ & $\times$ \\
CaP \cite{Liang2022CodeAP}    & $\checkmark$ & $\times$ & $\times$ & $\times$ & $\checkmark$ \\
MetaTool-ROS-Agent \textbf{(ours)}               & $\checkmark$ & $\checkmark$ & $\checkmark$ & $\checkmark$ & $\checkmark$ \\ 
\bottomrule
\end{tabularx}
\end{table*}

\section{Literature Review}
\label{sec:lit_review}
\noindent Recent advances in large language models (LLMs) have enabled new
robot control paradigms that integrate natural language reasoning with embodied
action. A representative example is NASA JPL's ROS-Agent, which bridges ROS-based
robots and language interfaces using LLMs to interpret user commands and invoke
ROS services through modular tool interfaces~\cite{royce2025enabling}. While
large closed-source models exhibit strong performance on complex tasks,
open-source LLMs often struggle with long-horizon planning unless augmented with
structured architectures or explicit memory mechanisms~\cite{xu2025alignment}.


To improve robustness, many systems augment LLMs with tools or structured
outputs. Toolformer trains LLMs to autonomously decide when and how to call
external APIs \cite{NEURIPS2023_d842425e}, while code-oriented models such as
Codex demonstrate that generating executable pseudo-code plans often referred
to as “code as policies” can effectively represent multi-step robotic behaviors
\cite{Liang2022CodeAP}. These approaches highlight the advantages of explicit,
program-like planning for long-horizon tasks.

A large body of recent work explores LLM-driven robotic agents that interleave
reasoning and acting or incorporate external tools and memory. ReAct prompts an
LLM to alternate between reasoning and action steps within a single loop
\cite{yao2022react}, while Zero-Shot Planners demonstrate that LLMs can decompose
high-level instructions into mid-level plans that are subsequently grounded to
robot actions \cite{pmlr-v162-huang22a}. Distinct from purely prompting-based planners, a recent direction fine-tunes open-source LLMs directly on embodied planning data: RoboGPT \cite{10891883} fine-tunes LLaMA on a large-scale embodied planning dataset to decompose natural-language instructions into long-horizon subgoal sequences, coupled with a re-planning module to correct infeasible plans. Other systems ground planning in richer
representations: SayPlan integrates hierarchical 3D scene graphs with classical
planning \cite{pmlr-v229-rana23a}, SPINE validates and refines subtasks in a
receding-horizon framework \cite{Ravichandran2024SPINEOS}, and ELLMER employs
retrieval-augmented prompting to generate feedback-aware, adaptive plans
\cite{Mon-Williams2025}.

Several works incorporate multimodal feedback or adaptive autonomy. Octopi
integrates tactile sensing into a vision-language model to support sequential
physical reasoning \cite{YuS-RSS-25}. CLEAR dynamically adjusts autonomy by
prompting an LLM to gate risky actions for human approval
\cite{10.1145/3610978.3640671}, while Matcha enables an LLM to iteratively select
epistemic actions and update plans using multimodal environmental feedback
\cite{Zhao2023ChatWT}.

Other approaches tightly couple LLMs with task-and-motion planning or specialized
control architectures. LLM3 iteratively refines symbolic and continuous plans
using motion planner feedback \cite{Wang2024LLM3LL}, FLTRNN introduces explicit
long- and short-term memory to maintain rule-consistent task execution
\cite{Zhang2024FLTRNNFL}, and CoPAL integrates layered reasoning, symbolic
planning, and motion generation with replanning for executability and safety
\cite{Joublin2023CoPALCP}. Additional systems include BTGenBot, which generates
ROS-compatible behavior trees \cite{Izzo2024BTGenBotBT}, SayCanPay, which combines
LLM proposals with affordance-aware cost evaluation
\cite{Hazra2023SayCanPayHP}, and LABOR, which coordinates bimanual manipulation
through explicit LLM planning \cite{Chu2024LargeLM}.

Finally, A3VLM explicitly reasons about object articulation and 
affordances~\cite{Huang2024A3VLMAA}, while Manipulate-Anything uses VLMs 
to autonomously plan, verify, and replan manipulation subgoals~\cite{pmlr-v270-duan25a}. 
Collectively, these systems demonstrate diverse agentic capabilities, yet 
long-horizon stability and execution reliability remain open challenges, 
particularly for open-source LLMs.

\section{Methodology}
\label{sec:methodology}
\noindent This section presents the proposed MetaTool-enhanced ROS-Agent architecture and describes its key components, including the agentic reasoning loop, action space, and memory mechanisms.

\subsection{Agentic Architecture, Action Space, and Memory}

\begin{figure*}[t]
    \centering
    \includegraphics[width=0.92\textwidth]{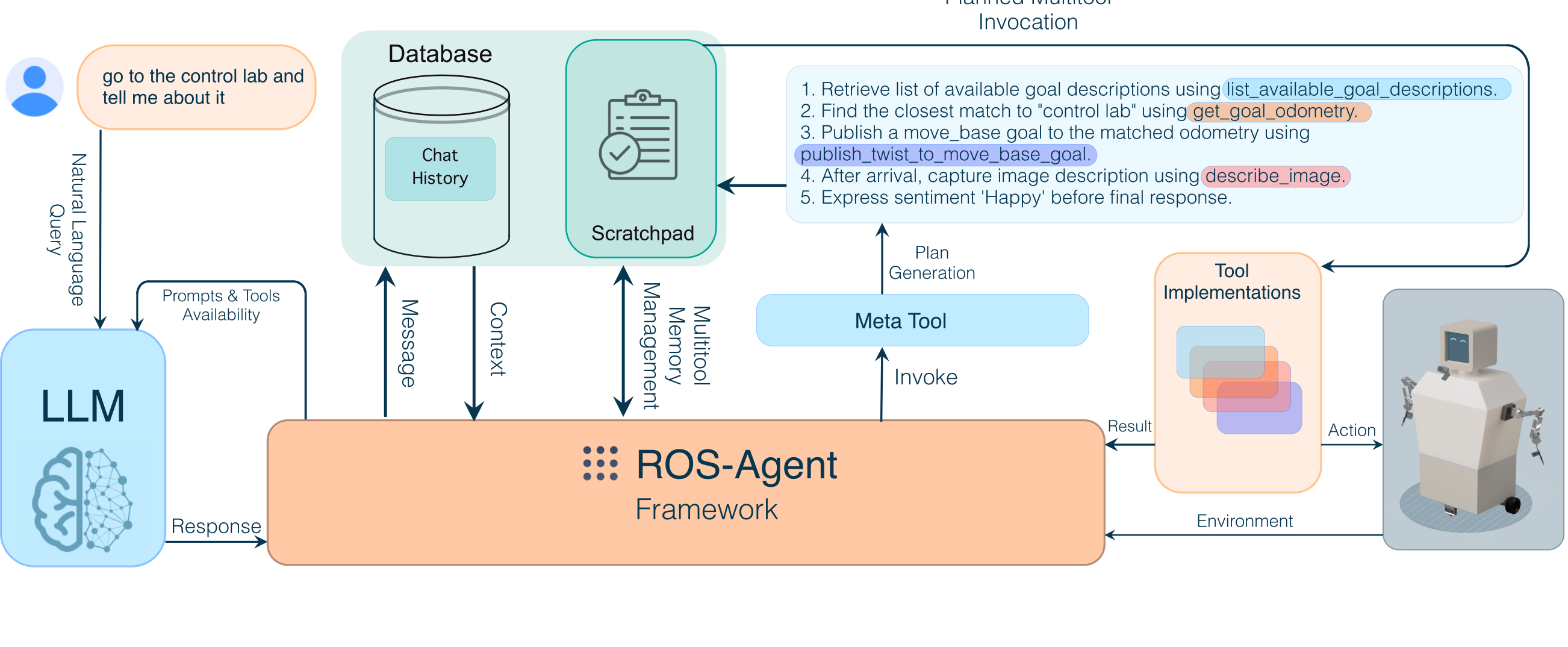}
    \caption{Functional architecture of the proposed MetaTool-enhanced ROS-Agentic framework.}
    \label{proposed_archi}
\end{figure*}

The proposed system adopts a Reasoning-and-Acting (ReAct) agentic paradigm
grounded in ROS-Agent's cognitive architecture \cite{royce2025enabling}, operating
through an iterative observation–reasoning–action loop. As shown in
 Fig.~\ref{proposed_archi}, a user query is forwarded to the LLM together with a
prefabricated system prompt that encodes the robot's embodiment and available
toolset, constraining the reasoning process to the platform's operational
capabilities. The LLM response is then passed to the ROS-Agent framework, which
bridges language-level intent and hardware execution through ROS.

Compared to baseline ROSA, the logic module is extended with a
MetaTool-driven planning stage that explicitly separates intent formation from
tool execution. Upon the first invocation, the MetaTool generates a structured
pseudo-code plan before any task-specific tool is executed. This plan is stored
in ROS-Agent's scratchpad, which is repurposed here as a structured execution buffer
that remains accessible throughout the entire interaction episode. Conditioning
each tool invocation on this externalized plan, rather than on transient local
context, stabilizes the execution trajectory and reduces repetitive or circular
tool calls that commonly occur in open-source LLMs during long-horizon,
multi-step tasks \cite{zhu2025llm}. 

The agent's action space includes ROS-Agent's native system tools
(e.g., \textit{rosnode}, \textit{rostopic}) and several custom task-oriented
modules for perception, affective expression, gesture control, and goal-directed
navigation. Examples include \textit{describe\_image} for vision–language
understanding, \textit{face\_expression} for social interaction, and
\textit{follow\_me} for human-following behaviors; additional implemented
tools and their functionalities are summarized in Table~\ref{tab:metatool_list}.

\begin{table}[]
\caption{Task-Oriented Tools Integrated into the MetaTool-Enhanced ROS Agent}
\label{tab:metatool_list}
\centering
\begin{tabular}{p{3.8cm} p{3.8cm}}
\toprule
\textbf{Tool} & \textbf{Functionality} \\
\midrule
\textit{describe\_image} & Vision–language understanding from live camera input \\
\textit{RAG\_query} & Knowledge retrieval via vector database querying \\
\textit{face\_expression} & Affective state publishing for social interaction \\
\textit{follow\_me} & Human-following behavior using pose keypoints \\
\textit{get\_goal\_odometry} & Goal-conditioned odometry estimation \\
\textit{set\_goal} & Goal definition and storage for navigation tasks \\
\textit{publish\_twist\_to\_move\_base} & High-level navigation goal dispatch \\
\textit{stop\_uh\_v1} & Emergency stop and goal cancellation \\
\bottomrule
\end{tabular}
\end{table}

\subsection{The MetaTool: Structured Reasoning and Execution}

The MetaTool represents the central architectural contribution of this work. It
functions as an intermediate agentic operator that separates task-level
planning from tool-level execution within the reasoning–action loop. Let
\( q \in \mathcal{Q} \) denote a natural-language user instruction and
\( \mathcal{A} = \{a_1, a_2, \dots, a_m\} \) the finite set of executable tools
available to the agent. The MetaTool first induces the language model to
generate an ordered symbolic plan

\[
P = \langle a_{i_1}, a_{i_2}, \dots, a_{i_n} \rangle, \quad a_{i_k} \in \mathcal{A},
\]

which decomposes \( q \) into a sequence of tool calls before any action is
executed. Unlike transient chain-of-thought traces, the plan \( P \) is
externalized into ROS-Agent's scratchpad memory and persists throughout the
observation–reasoning–action loop.

\begin{figure*}[]
    \centering
    \includegraphics[width=0.9\textwidth]{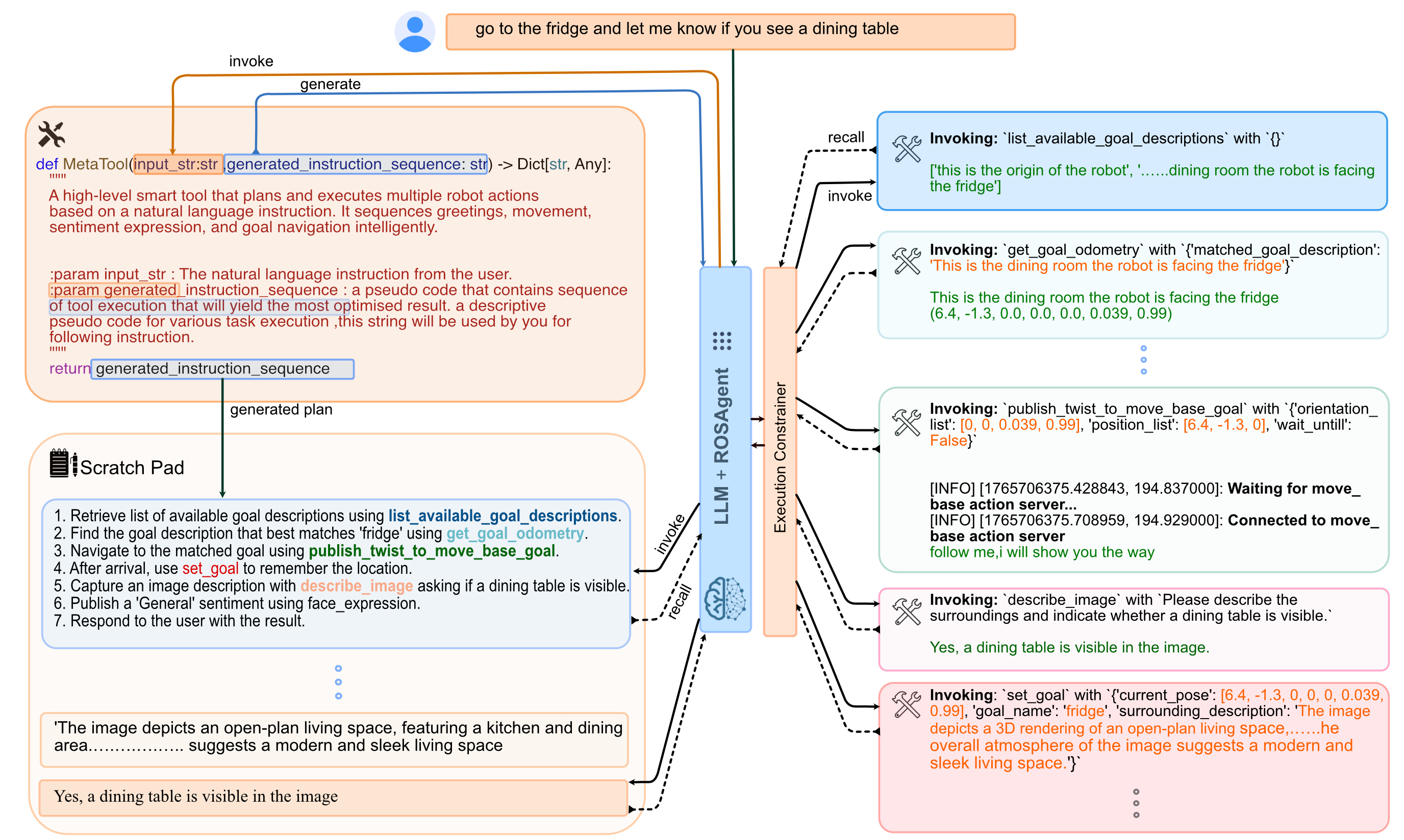}
    \caption{Operational overview of the proposed MetaTool integrated with the existing ROS agentic architecture, illustrating intermediate task planning and structured tool execution via scratchpad memory.}
    \label{meta_tool}
\end{figure*}

In a conventional agentic formulation, tool selection is governed by a policy
\( \pi(a_t \mid s_t) \), where \( s_t \in \mathcal{S} \) denotes the observable
state at step \( t \). The MetaTool replaces this with a plan-conditioned
execution policy

\[
\pi_{\mathrm{meta}}(a_t \mid s_t, P),
\]

which restricts admissible actions to those specified by the plan \( P \).
Formally,

\[
\pi_{\mathrm{meta}}(a_t \mid s_t, P) = 0 \quad \forall\, a_t \notin
\mathcal{A}_P(t),
\]

where \( \mathcal{A}_P(t) = \{a_{i_t}\} \) denotes the plan-specified tool at
step \( t \), unless explicit replanning is triggered. Since
\( |\mathcal{A}_P(t)| \ll |\mathcal{A}| \), this constraint reduces
unstructured action branching and prevents recursive loops and instruction
drift. As illustrated in Fig.~\ref{meta_tool}, the MetaTool (top-left) is invoked at
the start of each user query. It generates a structured pseudo-code plan,which is immediately stored in the scratchpad memory. This plan guides all subsequent multi-tool calls, with every tool's output appended to the scratchpad alongside the original plan. By continuously combining the initial plan with the evolving tool outputs (bottom-left of Fig.~\ref{meta_tool}), the scratchpad maintains a coherent execution trace that drives the tool execution chain consistently from start to finish.

The baseline ROS-Agent's scratchpad functions as an unconstrained conversational memory: it records what has occurred, but nothing prevents the agent's subsequent actions from diverging from its prior intent. MetaTool fundamentally redefines this memory as an enforceable execution contract. By formalizing tool selection as a plan-conditioned policy,
$\pi_{\mathrm{meta}}(a_t \mid s_t, P)$,
which effectively zeroes out the probability of any action outside the plan's admissible set,
$\mathcal{A}_P(t)$ especially for open-source LLMs, as demonstrated in the ablation studies, we transform planning from a passive record into an active constraint on the agent's action space. This reframing mitigates the long-horizon instability commonly observed in agentic systems. Furthermore, this plan-conditioned execution paradigm can, in principle, be adopted by similar frameworks as a drop-in mechanism, positioning MetaTool as an architectural primitive for constraining agentic reasoning under long-horizon uncertainty.

\subsection{Machine Configuration}
\noindent To evaluate the framework, we developed a cost-effective platform using an Intel-based onboard computer and an NVIDIA GTX 1050 Ti GPU. A custom motor driver fuses IMU and wheel encoder data for precise odometry. While the LLM operates via the Groq API \cite{groq2024api}, all perception and control models run locally on off-the-shelf hardware. The system architecture and interconnections are detailed in Fig.~\ref{overall_archi}.

\section{Evaluation}
\label{sec:evaluation}
\subsection{Experimental Setup and Metrics}
\label{subsec:exp_stp}
We evaluated the architecture using two environments: a ROS-Gazebo simulation and a custom physical mobile robot. The system was tested across three task levels ($T1$–$T3$) of increasing cognitive complexity, ranging from simple queries to contextually grounded sequential tasks.
\begin{itemize} 
\item \textbf{T1 (Simple Information Query):} Assesses basic knowledge retrieval and NLP parsing. \textit{Example: "Please brief me
about the operational ordinances of this facility. Are there any safety
regulations I should follow?"} 
\item \textbf{T2 (Complex Contextual Navigation):} Requires semantic reasoning to map high-level functional descriptions to specific locations. \textit{Example: "Is there a laboratory in this facility that conducts research on power system equipment? If yes, please lead me there."} 
\item \textbf{T3 (Contextually Grounded Sequential Task):} Evaluates long-horizon reasoning, human-following, and visual landmark persistence. \textit{Example: "Please follow me. [Robot follows the user.] Now observe and memorize this area [it includes a distinct red shelf. User introduces a unique landmark.] After some unrelated interactions, the user asks: ‘Can you show me the way to the red shelf I showed earlier?"} 
\end{itemize}
While $T1$ focuses on the 'Thought' aspect, tasks $T2$ and $T3$ are designed to evaluate the 'Action' and physical grounding of the MetaTool-enabled agent. We compared the proposed MetaTool-enhanced ROS Agent against the vanilla ROSA baseline \cite{royce2025enabling}. 
Evaluated models included proprietary LLMs and open-source models (e.g., GPT-OSS-20B, GPT-OSS-120B, Kimi-K2); the latter were selected based on demonstrated tool-invocation competence, evidenced by BFCL performance~\cite{patil2025bfcl} where available (Kimi-K2: 59.06\% overall, Rank~11) and by native function-calling API support and agentic benchmark results~\cite{agarwal2025gpt} for GPT-OSS variants. Collectively, these scores confirm that all three models meet a meaningful threshold of tool-use capability, forming a representative pool of open-source LLMs that spans both efficiency-optimised and reasoning-optimised deployment regimes.

\subsection{Ablation Study}

\begin{figure}[]
\centering
\includegraphics[width=0.48\textwidth]{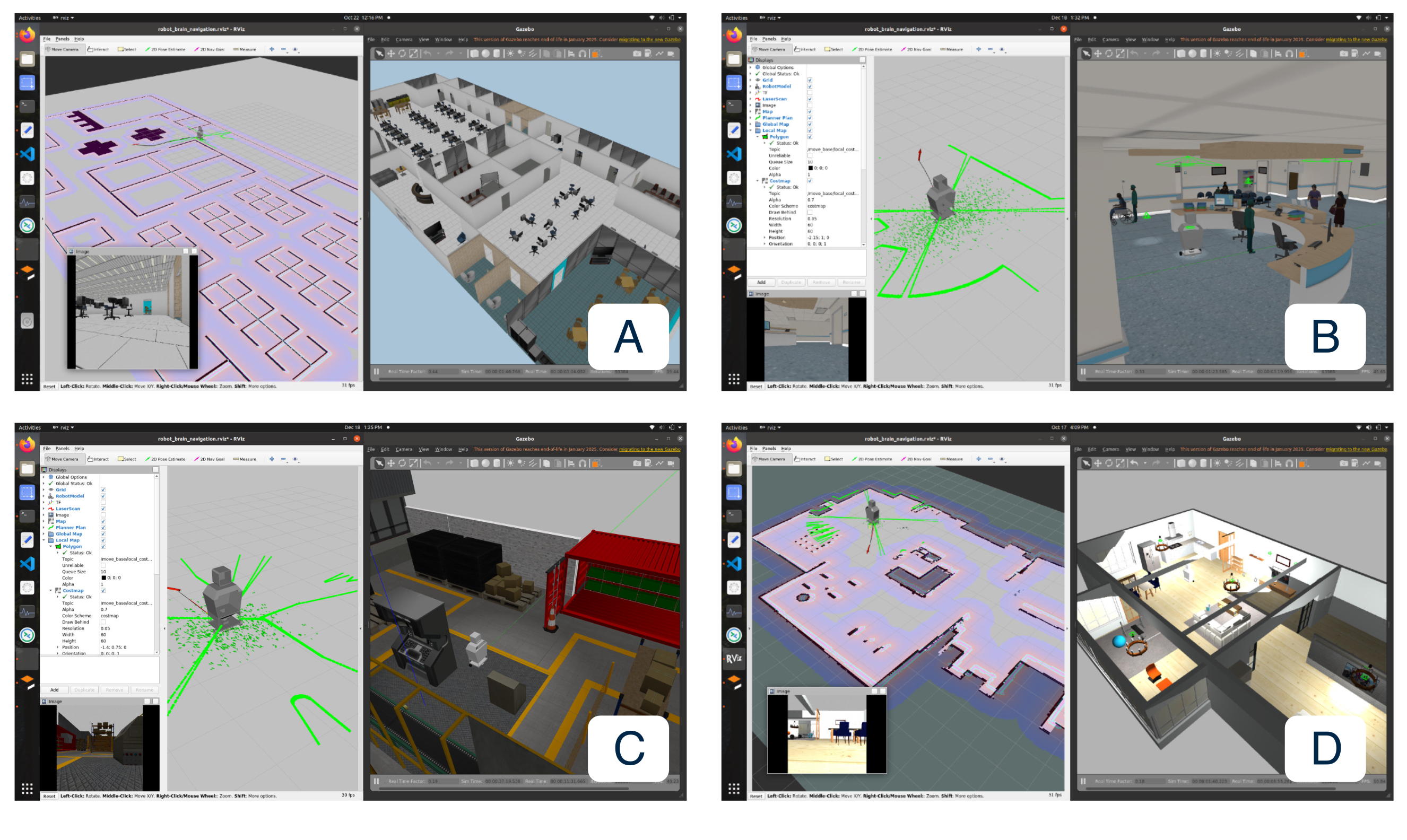}
\caption{Gazebo-based simulation and testing across diverse environments to ensure the robustness of agentic responses: (A) large office space; (B) hospital floor environment; (C) factory floor with scattered obstacles; and (D) standard urban home setting.}
\label{fig:sim_implementation}
\end{figure}

We conducted a component-level ablation study across four configurations ($C1$–$C4$) to isolate the contributions of persistent planning and execution constraints, as summarized in Table~\ref{tab:Ablation_comp_des}. To mitigate the inherent safety risks of autonomous LLM control on physical hardware, the ablation was performed within a Gazebo simulation environment (Fig.~\ref{fig:sim_implementation})\footnote{An anonymized codebase for the simulation and proposed execution pipeline is available at: \href{https://anonymous.4open.science/r/uh_bot_v1_sim-BDB1/README.md}{https://anonymous.4open.science/r/uh\_bot\_v1\_sim-BDB1/README.md}}. The agent was evaluated across diverse environments ,including a large office, a hospital floor, a factory floor with scattered obstacles, and a standard urban home to assess the robustness of its agentic responses. In this setup, $C1$ represents basic prompt engineering; $C2$ incorporates transient, non-persistent planning; $C3$ employs persistent scratchpad planning; and $C4$ (our proposed framework) combines ROS-Agent with MetaTool to enforce execution constraints by conditioning each tool invocation on a persistent task plan.

\begin{table*}[t]
\centering
\caption{Component-Wise Ablation Results Across Tasks. Task IDs are defined in Section~\ref{subsec:exp_stp}.}
\label{tab:Ablation}
\footnotesize
\begin{tabular}{@{}l p{6.5cm} lcccc@{}}
\toprule
\textbf{ID} & \textbf{Configuration} & \textbf{Task} &
\textbf{GPT-4o}~\cite{hurst2024gpt} &
\textbf{GPT-OSS-20B}~\cite{agarwal2025gpt} &
\textbf{GPT-OSS-120B}~\cite{agarwal2025gpt} &
\textbf{Kimi-K2}~\cite{team2025kimi} \\
\midrule

\multirow{3}{*}{C1} & \multirow{3}{=}{ROS-Agent + Adaptive Prompt} 
  & T1 & 15/20 (75\%) & 11/20 (55\%) & 11/20 (55\%) & 15/20 (75\%) \\
& & T2 & 10/20 (50\%) & 8/20 (40\%)  & 15/20 (75\%) & 17/20 (80\%) \\
& & T3 & 11/20 (55\%) & 6/20 (30\%)  & 8/20 (40\%)  & 10/20 (50\%) \\
\midrule

\multirow{3}{*}{C2} & \multirow{3}{=}{ROS-Agent + Prompt-Based Planner (Non-Persistent)} 
  & T1 & 17/20 (85\%) & 12/20 (60\%) & 15/20 (75\%) & 15/20 (75\%) \\
& & T2 & 17/20 (85\%) & 12/20 (60\%) & 15/20 (75\%) & 17/20 (80\%) \\
& & T3 & 15/20 (75\%) & 7/20 (35\%)  & 8/20 (40\%)  & 13/20 (65\%) \\
\midrule

\multirow{3}{*}{C3} & \multirow{3}{=}{ROS-Agent + MetaTool (Persistent Scratchpad, No Execution Constraint)} 
  & T1 & 19/20 (95\%) & 14/20 (70\%) & 18/20 (95\%) & 17/20 (85\%) \\
& & T2 & 17/20 (85\%) & 12/20 (60\%) & 16/20 (80\%) & 18/20 (85\%) \\
& & T3 & 15/20 (75\%) & 7/20 (35\%)  & 10/20 (50\%) & 13/20 (65\%) \\
\midrule

\multirow{3}{*}{C4} & \multirow{3}{=}{ROS-Agent + MetaTool (Persistent Scratchpad + Execution Constraint) \textbf{(Proposed)}} 
  & T1 & 18/20 (90\%) & 17/20 (85\%) & 17/20 (85\%) & 19/20 (95\%) \\
& & T2 & 17/20 (85\%) & 15/20 (75\%) & 18/20 (90\%) & 18/20 (85\%) \\
& & T3 & 15/20 (75\%) & 11/20 (55\%) & 14/20 (70\%) & 14/20 (70\%) \\
\bottomrule
\end{tabular}
\end{table*}

\begin{table}[!t]
\caption{MetaTool Component Ablation Configurations}
\label{tab:Ablation_comp_des}
\centering
\begin{tabular}{c p{3.5cm} p{3.9cm}}
\toprule
\textbf{ID} & \textbf{Configuration} & \textbf{Description} \\
\midrule
C1 & ROS-Agent + Adaptive Prompt &
Natural-language task execution guided solely by adaptive prompting, without explicit planning or persistent state. \\

C2 & ROS-Agent + Prompt-Based Planner (Non-Persistent) &
Prompt-based task decomposition generated at runtime; intermediate plans are not stored persistently and may be regenerated. \\

C3 & ROS-Agent + MetaTool (Persistent Scratchpad, No Execution Constraint) &
Explicit task plan stored in persistent scratchpad memory; tool selection is not constrained by the stored plan. \\

C4 & ROS-Agent + MetaTool (Persistent Scratchpad + Execution Constraint) &
Persistent task plan maintained and used to explicitly condition tool invocation, enforcing adherence throughout execution. \\
\bottomrule
\end{tabular}
\end{table}

Rather than comparing distinct architectures, all configurations are evaluated within the same agentic pipeline, differing only in the presence or absence of specific MetaTool components.Quantitative ablation results across tasks and language models are reported in Table~\ref{tab:Ablation}. For each configuration, task success is measured over 20 independent trials per task, and results are reported as empirical success rates. Due to the limited number of trials, these results are interpreted as indicative performance trends rather than statistically significant differences.
Key Findings and Analysis:
\begin{enumerate}

\item \textbf{Significance of Persistent Planning ($C1$ vs.\ $C3$):}
Transitioning from basic prompt-engineering ($C1$) to persistent scratchpad
planning ($C3$) yielded the most consistent performance gains across all LLMs.
For GPT-OSS-20B on T3, success rate improved from 30\% ($C1$) to 35\% ($C3$),
while GPT-OSS-120B improved from 40\% to 50\% on the same task demonstrating that persisting the MetaTool-generated pseudo-code in the scratchpad directly mitigates the context-forgetting behaviour common in open-source models during multi-step execution.

\item \textbf{Impact on Open-Source LLMs ($C2$ vs.\ $C3$):}
The benefit of persistent planning over transient prompt-based planning ($C2$)
was most pronounced for the open-source models. GPT-OSS-20B improved on T1
from 60\% ($C2$) to 70\% ($C3$), and Kimi-K2 improved on T1 from 75\% to
85\%. These gains reflect the role of explicit intent externalization in
reducing recursive tool invocation and premature action execution failure
modes that disproportionately affect smaller-scale models in ROS-native
environments.

\item \textbf{Execution Constraint Stability ($C3$ vs.\ $C4$):}
The addition of tool-level execution constraints in the proposed configuration
($C4$) provided measurable stability gains on the most demanding tasks. On T3,
GPT-OSS-20B improved from 35\% ($C3$) to 55\% ($C4$), GPT-OSS-120B from 50\%
to 70\%, and Kimi-K2 from 65\% to 70\%. On T2, GPT-OSS-120B reached 90\%
under $C4$ compared to 80\% under $C3$. These results confirm that strictly
constraining tool invocation to the MetaTool-prescribed plan is critical for
reliable convergence in long-horizon, multi-stage navigation tasks.

\end{enumerate}

The ablation results indicate that persistent scratchpad planning and execution constraints are key contributors to improved task success, with the most significant gains observed for open-source models on long-horizon tasks ($T2$, $T3$). Together, these components validate the core design rationale of the proposed MetaTool architecture. Among the open-source models tested, GPT-OSS-120B performed consistently better. Therefore, further evaluation with existing LLM-based robotic planning systems was conducted using GPT-OSS-120B as the LLM backbone.

\subsection{Overall Results and Efficiency Analysis}

\begin{figure*}[t]
\centering
\subfloat[]{\includegraphics[width=1.7in]{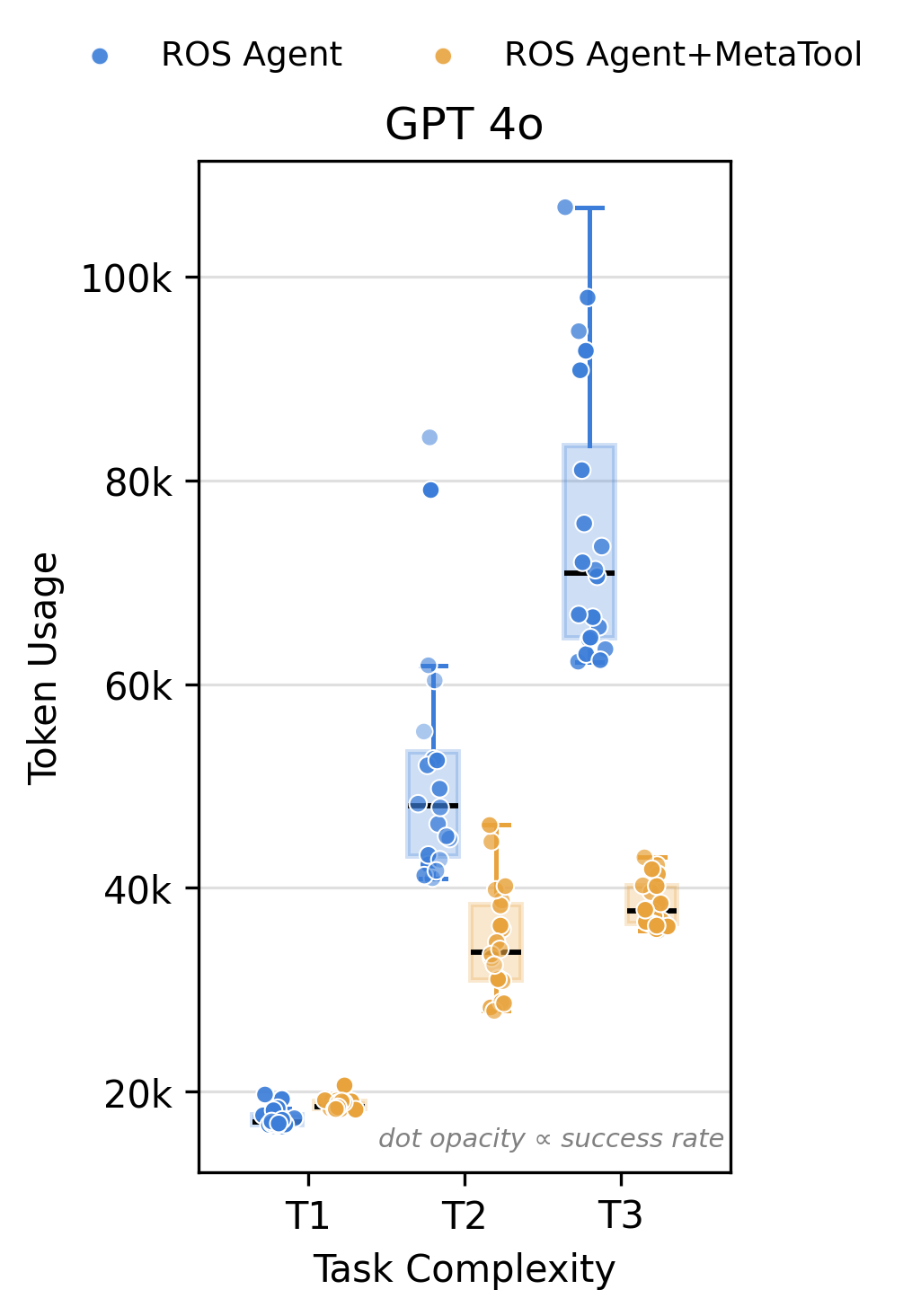}%
\label{a}}
\hfil
\subfloat[]{\includegraphics[width=1.7in]{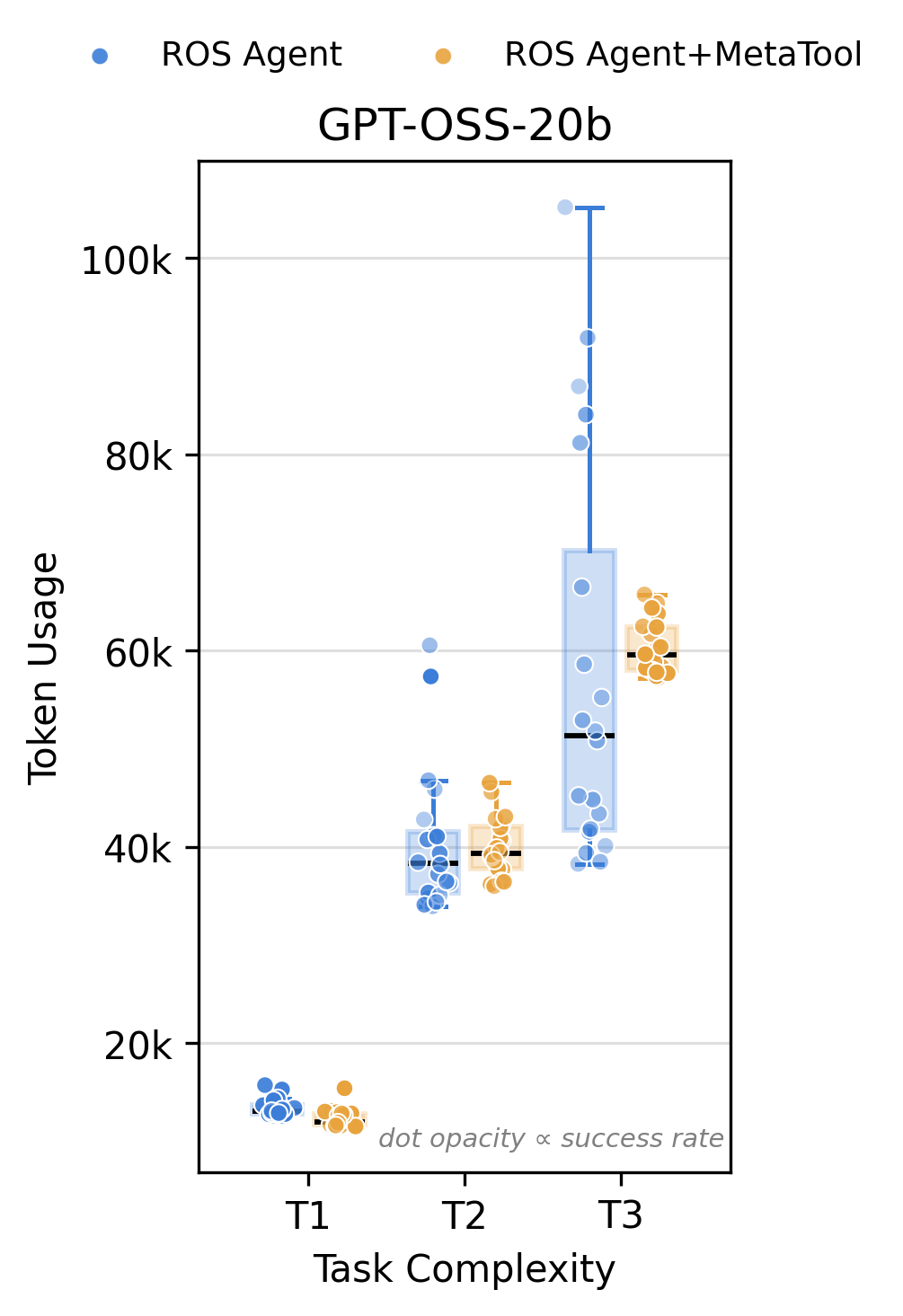}%
\label{b}}
\hfil
\subfloat[]{\includegraphics[width=1.7in]{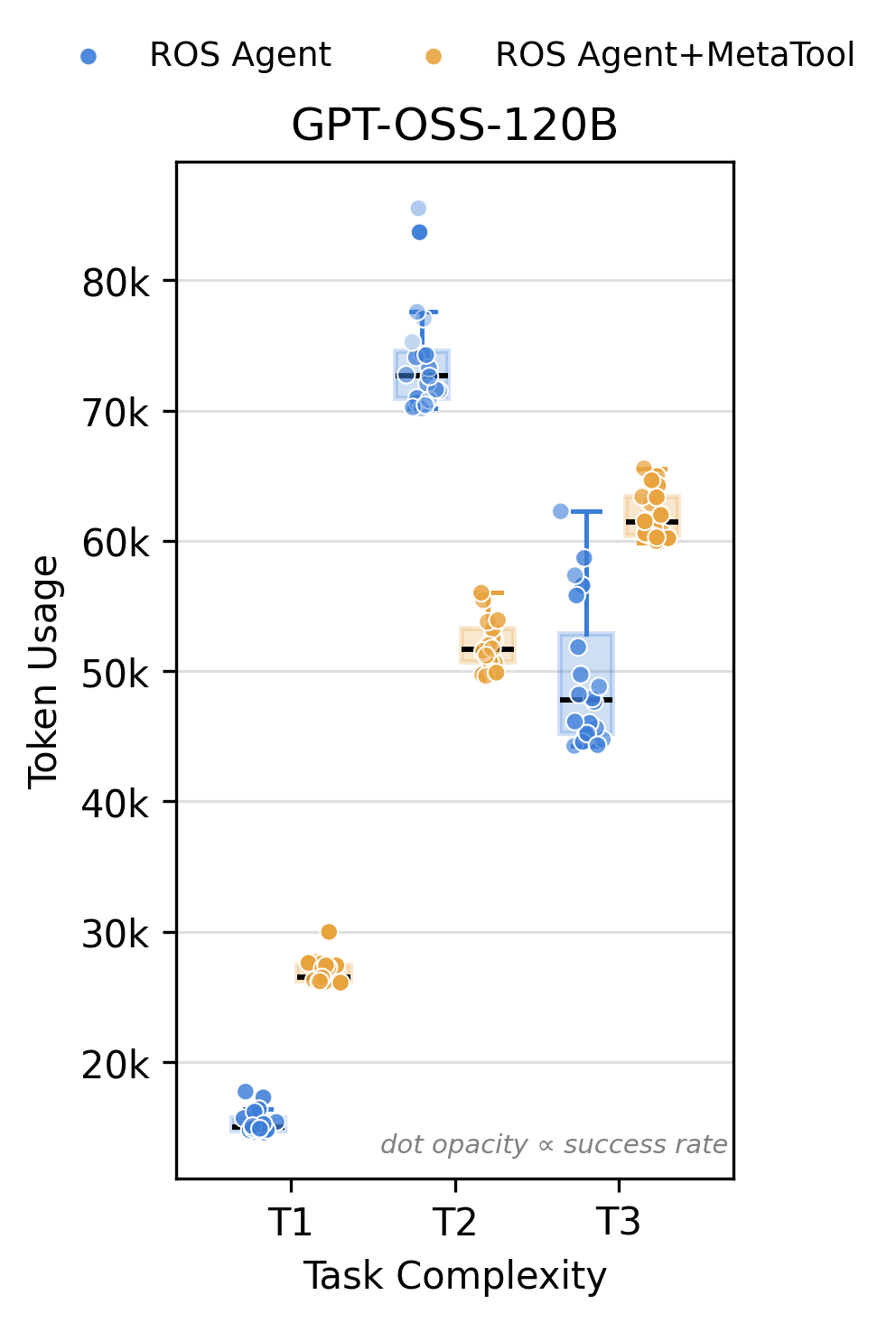}%
\label{c}}
\hfil
\subfloat[]{\includegraphics[width=1.7in]{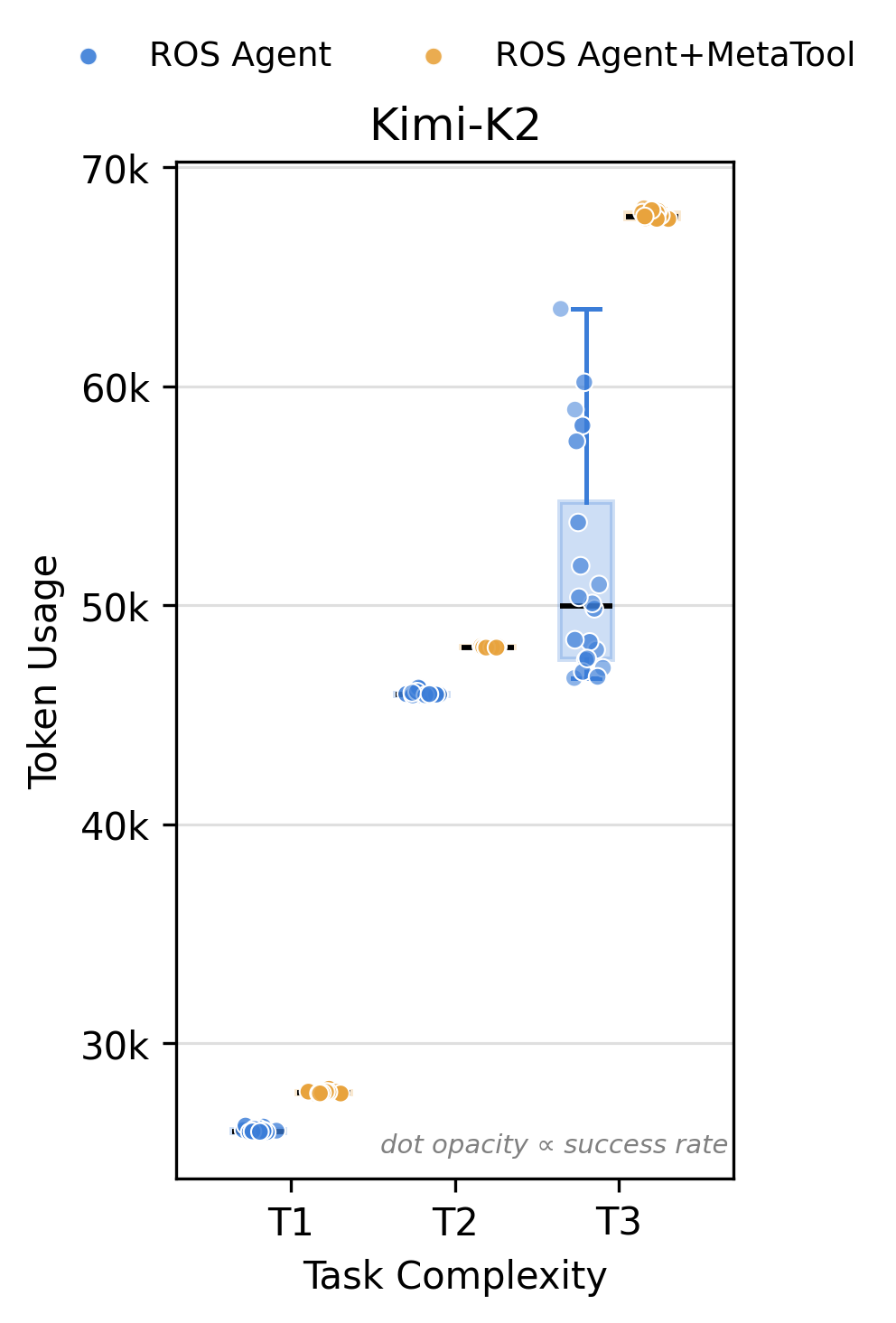}%
\label{d}}
\caption{Comparative Performance Analysis of Baseline ROSA vs. MetaTool-Enabled ROS Agent. The proposed framework demonstrates a higher Success Rate (indicated by the opacity) across all task complexities. The Token Usage Distribution highlights significantly lower variance in the proposed method, indicating that the MetaTool constraint prevents infinite loops and reasoning instability common in the open-source model.}
\label{token_usage}
\end{figure*}

The MetaTool-enhanced architecture demonstrated superior reasoning stability compared to vanilla ROS-Agent. Fig.~\ref{token_usage} compares vanilla ROS-Agent and MetaTool-enabled ROS-Agent across task complexities, highlighting success rates via marker opacity alongside token usage distributions. The high token variance observed in vanilla ROS-Agent indicates that identical queries frequently generate divergent tool sequences, reflecting unstable reasoning and inefficient convergence toward executable plans. In contrast, MetaTool produces compact, low-variance token profiles, demonstrating that it consistently generates structured, executable plans across repeated runs. This stabilizing effect is particularly pronounced in smaller open-source LLMs, where MetaTool effectively limits recursive tool invocations.

To rigorously substantiate these observations, we conducted an empirical evaluation across 40 independent trials on physical robotic hardware using the $T2$ task category. Evaluation was performed using the best performing open-source backbone, GPT-OSS-120B which demonstrated consistent performance among the evaluated open-source LLMs during the ablation study . We recorded average execution time alongside the frequency of repeated and invalid tool invocations.

\begin{table}[h]
\centering
\caption{Hardware evaluation of execution efficiency and stability across 40 trials using GPT-OSS-120B.All metrics are reported as mean $\pm$ standard deviation.}
\label{tab:Execution_Efficiency}
\begin{tabular}{@{}llll@{}}
\toprule
                                                              & \textbf{\begin{tabular}[c]{@{}l@{}}Repeated \\ Tool Calls (\%)\end{tabular}} & \textbf{\begin{tabular}[c]{@{}l@{}}Invalid/Failed \\ Tool Calls (\%)\end{tabular}} & \textbf{\begin{tabular}[c]{@{}l@{}}Execution \\ Time (s)\end{tabular}} \\ \midrule
ROS-Agent                                                     & 19.0 ± 5.7                                                              & 21.7 ± 6.5                                                                    & 89.51 ± 16.48                                                          \\
\begin{tabular}[c]{@{}l@{}}MetaTool (ours)\end{tabular} & 6.1 ± 2.2                                                               & 8.1 ± 2.9                                                                     & 84.23 ± 9.71                                                           \\ \bottomrule
\end{tabular}
\end{table}

As summarized in Table~\ref{tab:Execution_Efficiency}, both repeated and invalid tool invocations are reported as the mean $\pm$ standard deviation of their per-trial percentage relative to total tool calls, while execution time denotes task-completion latency in seconds across 40 real-world hardware trials. These empirical results align closely with the token-usage trends illustrated in Fig.~\ref{token_usage}, confirming that the reduced token variance of MetaTool directly translates to fewer redundant actions, decreased invocation failures, and more consistent execution latency in physical environments.

\begin{figure}[]
\centering
\includegraphics[width=0.48\textwidth]{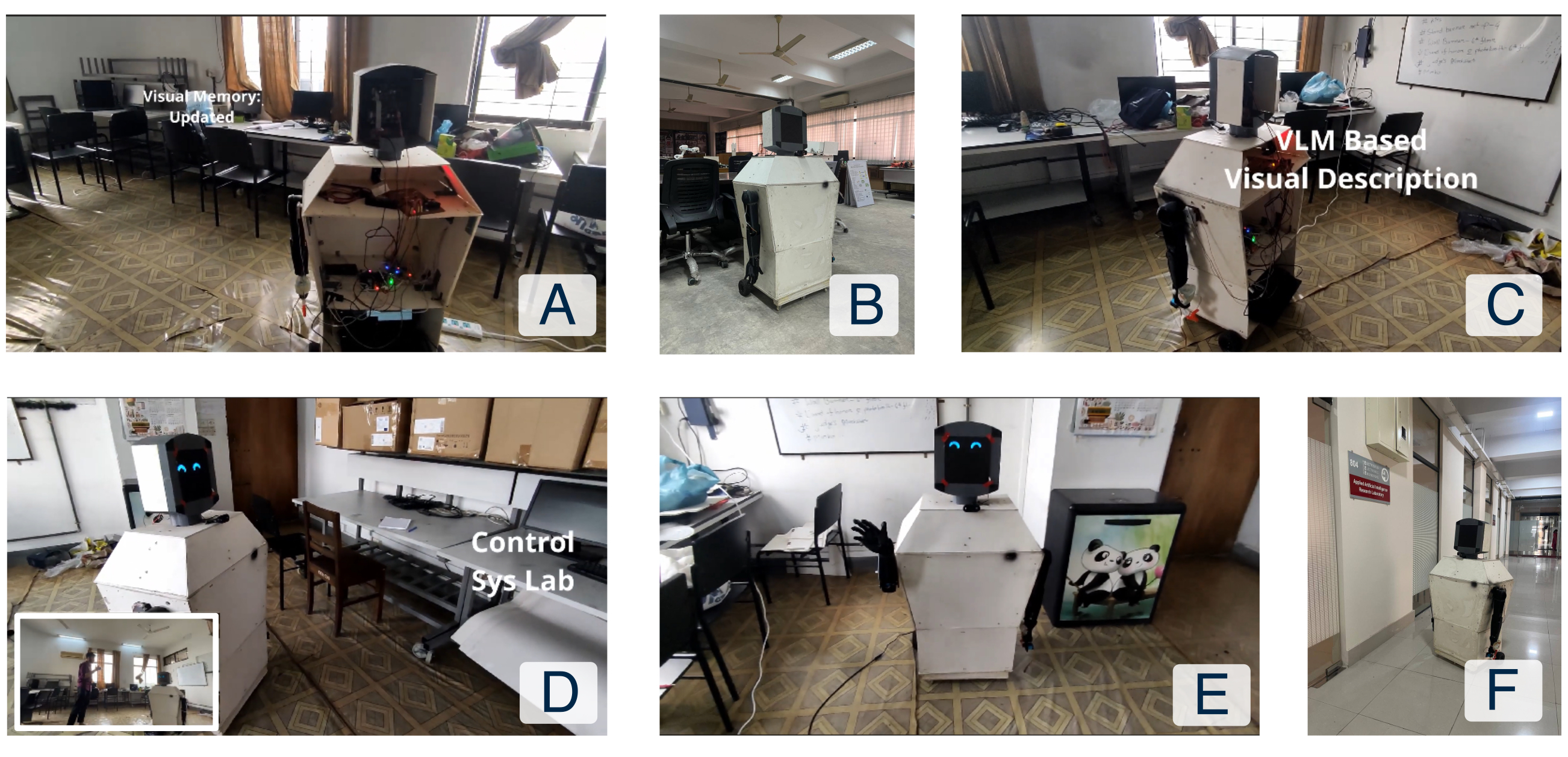}
\caption{Real-world implementation and experimental evaluation of the proposed architecture on a custom robotic platform: (A) vLM invocation for visual understanding; (B) autonomous navigation within an indoor laboratory environment; (C) visual memorization using agentic vLM calls for future spatial recollection;  (D) chauffeuring the user to a requested destination; (E) spontaneous user greeting based on predicted mood and perceived direction of communication; and (F) navigation through a narrow corridor while describing the surroundings to maintain user engagement.}
\label{fig:real_implementation}
\end{figure}

Finally, to evaluate the proposed framework against existing LLM-based planning paradigms, we conducted a controlled failure mode analysis on the physical robotic platform (shown in Fig.~\ref{fig:real_implementation}), comparing performance against re-implemented ReAct \cite{yao2022react} and Code-as-Policies (CaP) \cite{Liang2022CodeAP} baselines. Rather than using the original codebases directly, we re-implemented the core prompting strategy of each within our own agentic pipeline and ROS tool-calling interface. This adaptation was necessary because both paradigms were originally validated on environments incompatible with our platform. ReAct on text-based QA and interactive text environments (HotpotQA, ALFWorld), and CaP on tabletop manipulation in simulation with a distinct action/perception API. This re-implementation preserves each method's defining mechanism (interleaved reasoning-action with no persistent plan buffer for ReAct and single-shot plan generation without execution constraints for CaP) while grounding them in our specific tool set, task suite ($T1$–$T3$), and physical robot, thereby enabling a direct, controlled comparison against the proposed MetaTool-ROSA architecture under identical conditions.

\begin{table}[]
\centering
\caption{Quantitative comparison of failure modes and overall task success rates across evaluated planning frameworks.}
\label{tab:failure_mode}
\begin{tabular}{@{}lllll@{}}
\toprule
\textbf{Method}                                                     & \textbf{\begin{tabular}[c]{@{}l@{}}Incorrect / \\ Hallucinated\\  Plan\end{tabular}} & \textbf{\begin{tabular}[c]{@{}l@{}}Navigation \\ Failure\end{tabular}} & \textbf{\begin{tabular}[c]{@{}l@{}}Tool \\ Unavailable / \\ Timeout\end{tabular}} & \textbf{\begin{tabular}[c]{@{}l@{}}Successful \\ Trials\end{tabular}} \\ \midrule
\textbf{ReAct \cite{yao2022react}}                                                      & 17/40                                                                                & 13/40                                                                  & 10/40                                                                             & 17/40                                                                 \\
\textbf{CaP \cite{Liang2022CodeAP}}                                                        & 13/40                                                                                & 10/40                                                                  & 13/40                                                                             & 20/40                                                                 \\
\textbf{ROSA \cite{royce2025enabling}}                                                       & 17/40                                                                                & 10/40                                                                  & 17/40                                                                             & 23/40                                                                 \\
\textbf{\begin{tabular}[c]{@{}l@{}}MetaTool \\ (ours)\end{tabular}} & 10/40                                                                                & 7/40                                                                   & 3/40                                                                              & 30/40                                                                 \\ \bottomrule
\multicolumn{5}{p{0.96\columnwidth}}{\vspace{2pt}\footnotesize \textit{Note:} Failure categories are non-exclusive. Individual failed trials may encounter multiple failure modes, so total category counts may exceed total failed trials.}
\end{tabular}
\end{table}

As detailed in Table~\ref{tab:failure_mode}, the absence of a persistent plan buffer in the ReAct baseline exacerbates recursive reasoning degradation during long-horizon tasks, resulting in a high incidence of incorrect or hallucinated plans (17/40) and navigation failures (13/40), culminating in a low overall success rate of 42.5\% (17/40). The CaP-style planner slightly mitigates planning hallucinations (13/40) but remains highly susceptible to tool unavailability and execution timeouts (13/40), limiting its task success to 50.0\% (20/40). While the vanilla ROSA pipeline demonstrates a marginal improvement with a 57.5\% success rate (23/40 successful trials), it still exhibits significant vulnerability to unconstrained tool invocations, reflected by high rates of hallucinated plans (17/40) and tool timeouts (17/40). Conversely, our proposed MetaTool framework significantly reduces these failure modes across all categories most notably dropping tool unavailability/timeout to just 3/40 achieving a superior overall success rate of 75.0\% (30/40). These empirical findings confirm that unconstrained, prompt-based reasoning mechanisms struggle with the stability demands of real-world robotic deployment. Consequently, the observed failure modes directly underscore the necessity of the persistent scratchpad and structural execution constraints introduced in the MetaTool-ROS-Agent architecture, which successfully translate simulation-based efficiency gains into robust physical operation.

\section{Conclusion and Future Work}
\label{sec:conclusion}

\noindent This work presented a MetaTool enhanced ROS-Agent architecture for
structured agentic task execution in LLM-driven robotic systems. By enforcing
an explicit planning stage that generates a pseudo-code task sequence prior to
tool invocation, the proposed approach separates high-level reasoning from
execution within the ROS-Agent framework. Experimental results demonstrate improved
reasoning stability, task coordination, and execution reliability for
open-source LLMs, particularly in complex multi-step tasks.

Future work will investigate adapting the architecture to smaller open-source
language models to enable efficient deployment on resource-constrained robotic
platforms. In addition, we plan to extend the robotic platform and tool
ecosystem to support broader real-world applications of structured agentic
robotic systems.


\bibliographystyle{IEEEtran}
\bibliography{main}

@INPROCEEDINGS{10730448,
  author={Raja, Abhinav and Bhethanabotla, Anish},
  booktitle={2024 IEEE 1st International Conference on Communication Engineering and Emerging Technologies (ICoCET)}, 
  title={OperateLLM: Integrating Robot Operating System (ROS) Tools in Large Language Models}, 
  year={2024},
  volume={},
  number={},
  pages={1-4},
  doi={10.1109/ICoCET63343.2024.10730448}}

@ARTICLE{10891883,
  author={Chen, Yaran and Cui, Wenbo and Chen, Yuanwen and Tan, Mining and Zhang, Xinyao and Liu, Jinrui and Li, Haoran and Zhao, Dongbin and Wang, He},
  journal={IEEE Transactions on Cognitive and Developmental Systems}, 
  title={RoboGPT: An LLM-Based Long-Term Decision-Making Embodied Agent for Instruction Following Tasks}, 
  year={2025},
  volume={17},
  number={5},
  pages={1163-1174},
  doi={10.1109/TCDS.2025.3543364}}

@INPROCEEDINGS{10610981,
  author={Sun, Lingfeng and Jha, Devesh K. and Hori, Chiori and Jain, Siddarth and Corcodel, Radu and Zhu, Xinghao and Tomizuka, Masayoshi and Romeres, Diego},
  booktitle={2024 IEEE International Conference on Robotics and Automation (ICRA)}, 
  title={Interactive Planning Using Large Language Models for Partially Observable Robotic Tasks}, 
  year={2024},
  volume={},
  number={},
  pages={14054-14061},
  doi={10.1109/ICRA57147.2024.10610981}}

@inproceedings{royce2025enabling,
  title={Enabling novel mission operations and interactions with rosa: The robot operating system agent},
  author={Royce, Rob and Kaufmann, Marcel and Becktor, Jonathan and Moon, Sangwoo and Carpenter, Kalind and Pak, Kai and Towler, Amanda and Thakker, Rohan and Khattak, Shehryar},
  booktitle={2025 IEEE Aerospace Conference},
  pages={1--16},
  year={2025},
  organization={IEEE}
}

@inproceedings{patil2025bfcl,
title={The Berkeley Function Calling Leaderboard (BFCL): From Tool Use to Agentic Evaluation of Large Language Models}, 
author={Patil, Shishir G. and Mao, Huanzhi and Cheng-Jie Ji, Charlie and Yan, Fanjia and Suresh, Vishnu and Stoica, Ion and E. Gonzalez, Joseph},
booktitle={Forty-second International Conference on Machine Learning},
year={2025},
}

@inproceedings{Quigley2009ROSAO,
  title={ROS: an open-source Robot Operating System},
  author={Morgan Quigley},
  booktitle={IEEE International Conference on Robotics and Automation},
  year={2009},
  url={https://api.semanticscholar.org/CorpusID:6324125}
}

@ARTICLE{6174325,
  author={Chitta, Sachin and Sucan, Ioan and Cousins, Steve},
  journal={IEEE Robotics \& Automation Magazine}, 
  title={MoveIt! [ROS Topics]}, 
  year={2012},
  volume={19},
  number={1},
  pages={18-19},
  doi={10.1109/MRA.2011.2181749}}

@article{Liu2023LostIT,
  title={Lost in the Middle: How Language Models Use Long Contexts},
  author={Nelson F. Liu and Kevin Lin and John Hewitt and Ashwin Paranjape and Michele Bevilacqua and Fabio Petroni and Percy Liang},
  journal={Transactions of the Association for Computational Linguistics},
  year={2023},
  volume={12},
  pages={157-173},
  url={https://api.semanticscholar.org/CorpusID:259360665}
}

@inproceedings{xu2025alignment,
  title={Alignment for efficient tool calling of large language models},
  author={Xu, Hongshen and Wang, Zihan and Zhu, Zichen and Pan, Lei and Chen, Xingyu and Fan, Shuai and Chen, Lu and Yu, Kai},
  booktitle={Proceedings of the 2025 Conference on Empirical Methods in Natural Language Processing},
  pages={17787--17803},
  year={2025}
}

@inproceedings{NEURIPS2023_d842425e,
 author = {Schick, Timo and Dwivedi-Yu, Jane and Dessi, Roberto and Raileanu, Roberta and Lomeli, Maria and Hambro, Eric and Zettlemoyer, Luke and Cancedda, Nicola and Scialom, Thomas},
 booktitle = {Advances in Neural Information Processing Systems},
 editor = {A. Oh and T. Naumann and A. Globerson and K. Saenko and M. Hardt and S. Levine},
 pages = {68539--68551},
 publisher = {Curran Associates, Inc.},
 title = {Toolformer: Language Models Can Teach Themselves to Use Tools},
 url = {https://proceedings.neurips.cc/paper_files/paper/2023/file/d842425e4bf79ba039352da0f658a906-Paper-Conference.pdf},
 volume = {36},
 year = {2023}
}

@INPROCEEDINGS{10610065,
  author={Zhou, Zhehua and Song, Jiayang and Yao, Kunpeng and Shu, Zhan and Ma, Lei},
  booktitle={2024 IEEE International Conference on Robotics and Automation (ICRA)}, 
  title={ISR-LLM: Iterative Self-Refined Large Language Model for Long-Horizon Sequential Task Planning}, 
  year={2024},
  volume={},
  number={},
  pages={2081-2088},
  doi={10.1109/ICRA57147.2024.10610065}}

@INPROCEEDINGS{1389727,
  author={Koenig, N. and Howard, A.},
  booktitle={2004 IEEE/RSJ International Conference on Intelligent Robots and Systems (IROS) (IEEE Cat. No.04CH37566)}, 
  title={Design and use paradigms for Gazebo, an open-source multi-robot simulator}, 
  year={2004},
  volume={3},
  number={},
  pages={2149-2154 vol.3},
  doi={10.1109/IROS.2004.1389727}}

@article{Liang2022CodeAP,
  title={Code as Policies: Language Model Programs for Embodied Control},
  author={Jacky Liang and Wenlong Huang and F. Xia and Peng Xu and Karol Hausman and Brian Ichter and Peter R. Florence and Andy Zeng},
  journal={2023 IEEE International Conference on Robotics and Automation (ICRA)},
  year={2022},
  pages={9493-9500},
  url={https://api.semanticscholar.org/CorpusID:252355542}
}

@article{yao2022react,
  title={ReAct: Synergizing Reasoning and Acting in Language Models},
  author={Yao, Shunyu and Zhao, Jeffrey and Yu, Dian and Du, Nan and Shafran, Izhak and Narasimhan, Karthik and Cao, Yuan},
  journal={arXiv preprint arXiv:2210.03629},
  year={2022}
}

@InProceedings{pmlr-v162-huang22a,
  title = 	 {Language Models as Zero-Shot Planners: Extracting Actionable Knowledge for Embodied Agents},
  author =       {Huang, Wenlong and Abbeel, Pieter and Pathak, Deepak and Mordatch, Igor},
  booktitle = 	 {Proceedings of the 39th International Conference on Machine Learning},
  pages = 	 {9118--9147},
  year = 	 {2022},
  editor = 	 {Chaudhuri, Kamalika and Jegelka, Stefanie and Song, Le and Szepesvari, Csaba and Niu, Gang and Sabato, Sivan},
  volume = 	 {162},
  series = 	 {Proceedings of Machine Learning Research},
  month = 	 {17--23 Jul},
  publisher =    {PMLR},
  url = 	 {https://proceedings.mlr.press/v162/huang22a.html}
}

@InProceedings{pmlr-v229-rana23a,
  title = 	 {SayPlan: Grounding Large Language Models using 3D Scene Graphs for Scalable Robot Task Planning},
  author =       {Rana, Krishan and Haviland, Jesse and Garg, Sourav and Abou-Chakra, Jad and Reid, Ian and Suenderhauf, Niko},
  booktitle = 	 {Proceedings of The 7th Conference on Robot Learning},
  pages = 	 {23--72},
  year = 	 {2023},
  editor = 	 {Tan, Jie and Toussaint, Marc and Darvish, Kourosh},
  volume = 	 {229},
  series = 	 {Proceedings of Machine Learning Research},
  month = 	 {06--09 Nov},
  publisher =    {PMLR},
  url = 	 {https://proceedings.mlr.press/v229/rana23a.html}
}

@article{Ravichandran2024SPINEOS,
  title={SPINE: Online Semantic Planning for Missions with Incomplete Natural Language Specifications in Unstructured Environments},
  author={Zachary Ravichandran and Varun Murali and Mariliza Tzes and George J. Pappas and Vijay Kumar},
  journal={2025 IEEE International Conference on Robotics and Automation (ICRA)},
  year={2024},
  pages={13714-13721},
  url={https://api.semanticscholar.org/CorpusID:273162481}
}

@article{Mon-Williams2025,
  author    = {Ruaridh Mon-Williams and Gen Li and Ran Long and Wenqian Du and Christopher G. Lucas},
  title     = {Embodied large language models enable robots to complete complex tasks in unpredictable environments},
  journal   = {Nature Machine Intelligence},
  year      = {2025},
  volume    = {7},
  number    = {4},
  pages     = {592--601},
  doi       = {10.1038/s42256-025-01005-x},
  url       = {https://doi.org/10.1038/s42256-025-01005-x},
  issn      = {2522-5839}
}

@INPROCEEDINGS{YuS-RSS-25,
AUTHOR = {Samson Yu AND Lin Kelvin AND Harold Soh},
TITLE = {{Demonstrating the Octopi-1.5 Visual-Tactile-Language Model}},
BOOKTITLE = {Proceedings of Robotics: Science and Systems},
YEAR = {2025},
ADDRESS = {LosAngeles, CA, USA},
MONTH = {June},
DOI = {10.15607/RSS.2025.XXI.058}
}

@inproceedings{10.1145/3610978.3640671,
author = {Macdonald, Jacob P. and Mallick, Rohit and Wollaber, Allan B. and Pe\~{n}a, Jaime D. and McNeese, Nathan and Siu, Ho Chit},
title = {Language, Camera, Autonomy! Prompt-engineered Robot Control for Rapidly Evolving Deployment},
year = {2024},
isbn = {9798400703232},
publisher = {Association for Computing Machinery},
address = {New York, NY, USA},
url = {https://doi.org/10.1145/3610978.3640671},
doi = {10.1145/3610978.3640671},
booktitle = {Companion of the 2024 ACM/IEEE International Conference on Human-Robot Interaction},
pages = {717–721},
numpages = {5},
location = {Boulder, CO, USA},
series = {HRI '24}
}

@article{Zhao2023ChatWT,
  title={Chat with the Environment: Interactive Multimodal Perception Using Large Language Models},
  author={Xufeng Zhao and Mengdi Li and Cornelius Weber and Muhammad Burhan Hafez and Stefan Wermter},
  journal={2023 IEEE/RSJ International Conference on Intelligent Robots and Systems (IROS)},
  year={2023},
  pages={3590-3596},
  url={https://api.semanticscholar.org/CorpusID:257532768}
}

@article{zhu2025llm,
  title={Where llm agents fail and how they can learn from failures},
  author={Zhu, Kunlun and Liu, Zijia and Li, Bingxuan and Tian, Muxin and Yang, Yingxuan and Zhang, Jiaxun and Han, Pengrui and Xie, Qipeng and Cui, Fuyang and Zhang, Weijia and others},
  journal={arXiv preprint arXiv:2509.25370},
  year={2025}
}

@INPROCEEDINGS{Wang2024LLM3LL,
  author={Wang, Shu and Han, Muzhi and Jiao, Ziyuan and Zhang, Zeyu and Wu, Ying Nian and Zhu, Song-Chun and Liu, Hangxin},
  booktitle={2024 IEEE/RSJ International Conference on Intelligent Robots and Systems (IROS)}, 
  title={LLM3: Large Language Model-based Task and Motion Planning with Motion Failure Reasoning}, 
  year={2024},
  volume={},
  number={},
  pages={12086-12092},
  doi={10.1109/IROS58592.2024.10801328}}

@article{Zhang2024FLTRNNFL,
  title={FLTRNN: Faithful Long-Horizon Task Planning for Robotics with Large Language Models},
  author={Jiatao Zhang and LanLing Tang and Yufan Song and Qiwei Meng and Haofu Qian and Jun Shao and Wei Song and Shiqiang Zhu and Jason Jianjun Gu},
  journal={2024 IEEE International Conference on Robotics and Automation (ICRA)},
  year={2024},
  pages={6680-6686},
  url={https://api.semanticscholar.org/CorpusID:271798999}
}

@article{Joublin2023CoPALCP,
  title={CoPAL: Corrective Planning of Robot Actions with Large Language Models},
  author={Frank Joublin and Antonello Ceravola and Pavel Smirnov and Felix Ocker and Joerg Deigmoeller and Anna Belardinelli and Chao Wang and Stephan Hasler and Daniel Tanneberg and Michael Gienger},
  journal={2024 IEEE International Conference on Robotics and Automation (ICRA)},
  year={2023},
  pages={8664-8670},
  url={https://api.semanticscholar.org/CorpusID:263835008}
}

@article{Izzo2024BTGenBotBT,
  title={BTGenBot: Behavior Tree Generation for Robotic Tasks with Lightweight LLMs},
  author={Riccardo Andrea Izzo and Gianluca Bardaro and Matteo Matteucci},
  journal={2024 IEEE/RSJ International Conference on Intelligent Robots and Systems (IROS)},
  year={2024},
  pages={9684-9690},
  url={https://api.semanticscholar.org/CorpusID:268531996}
}

@article{Hazra2023SayCanPayHP,
  title={SayCanPay: Heuristic Planning with Large Language Models using Learnable Domain Knowledge},
  author={Rishi Hazra and Pedro Zuidberg Dos Martires and Luc De Raedt},
  journal={ArXiv},
  year={2023},
  volume={abs/2308.12682},
  url={https://api.semanticscholar.org/CorpusID:261100610}
}

@article{Chu2024LargeLM,
  title={Large Language Models for Orchestrating Bimanual Robots},
  author={Kun Chu and Xufeng Zhao and Cornelius Weber and Mengdi Li and Wenhao Lu and Stefan Wermter},
  journal={2024 IEEE-RAS 23rd International Conference on Humanoid Robots (Humanoids)},
  year={2024},
  pages={328-334},
  url={https://api.semanticscholar.org/CorpusID:268856981}
}

@article{Huang2024A3VLMAA,
  title={A3VLM: Actionable Articulation-Aware Vision Language Model},
  author={Siyuan Huang and Haonan Chang and Yuhan Liu and Yimeng Zhu and Hao Dong and Peng Gao and Abdeslam Boularias and Hongsheng Li},
  journal={ArXiv},
  year={2024},
  volume={abs/2406.07549},
  url={https://api.semanticscholar.org/CorpusID:270379695}
}

@InProceedings{pmlr-v270-duan25a,
  title = 	 {Manipulate-Anything: Automating Real-World Robots using Vision-Language Models},
  author =       {Duan, Jiafei and Yuan, Wentao and Pumacay, Wilbert and Wang, Yi Ru and Ehsani, Kiana and Fox, Dieter and Krishna, Ranjay},
  booktitle = 	 {Proceedings of The 8th Conference on Robot Learning},
  pages = 	 {5326--5350},
  year = 	 {2025},
  editor = 	 {Agrawal, Pulkit and Kroemer, Oliver and Burgard, Wolfram},
  volume = 	 {270},
  series = 	 {Proceedings of Machine Learning Research},
  month = 	 {06--09 Nov},
  publisher =    {PMLR},
  url = 	 {https://proceedings.mlr.press/v270/duan25a.html}
}

@article{agarwal2025gpt,
  title={gpt-oss-120b \& gpt-oss-20b model card},
  author={Agarwal, Sandhini and Ahmad, Lama and Ai, Jason and Altman, Sam and Applebaum, Andy and Arbus, Edwin and Arora, Rahul K and Bai, Yu and Baker, Bowen and Bao, Haiming and others},
  journal={arXiv preprint arXiv:2508.10925},
  year={2025}
}

@article{team2025kimi,
  title={Kimi k2: Open agentic intelligence},
  author={Team, Kimi and Bai, Yifan and Bao, Yiping and Chen, Guanduo and Chen, Jiahao and Chen, Ningxin and Chen, Ruijue and Chen, Yanru and Chen, Yuankun and Chen, Yutian and others},
  journal={arXiv preprint arXiv:2507.20534},
  year={2025}
}

@article{hurst2024gpt,
  title={Gpt-4o system card},
  author={Hurst, Aaron and Lerer, Adam and Goucher, Adam P and Perelman, Adam and Ramesh, Aditya and Clark, Aidan and Ostrow, AJ and Welihinda, Akila and Hayes, Alan and Radford, Alec and others},
  journal={arXiv preprint arXiv:2410.21276},
  year={2024}
}

@misc{groq2024api,
  author = {{Groq Team}},
  title = {GroqCloud API Documentation},
  year = {2024},
  publisher = {Groq},
  url = {https://console.groq.com/docs/},
  note = {Accessed: 2026-01-09}
}


\end{document}